\documentclass[10pt,twocolumn,letterpaper]{article}

\usepackage{wacv}
\usepackage{times}
\usepackage{epsfig}
\usepackage{graphicx}
\usepackage{amsmath}
\usepackage{amssymb}

\def\wacvPaperID{656} 

\wacvfinalcopy 

\ifwacvfinal
\def\assignedStartPage{9876} 
\fi

\ifwacvfinal
\usepackage[breaklinks=true,bookmarks=false]{hyperref}
\else
\usepackage[pagebackref=true,breaklinks=true,colorlinks,bookmarks=false]{hyperref}
\fi

\ifwacvfinal
\else
\fi

\begin{document}

\title{Egocentric Hand-object Interaction Detection and Application}

\author{Yao LU\\
Bristol University\\
{\tt\small yl1220@brsitol.ac.uk}
\and
Walterio W. Mayol-Cuevas\\
Bristol University\\
{\tt\small wmayol@cs.bris.ac.uk}
}

\maketitle

\begin{abstract}
In this paper, we present a method to detect the hand-object interaction from egocentric perspective. In contrast to massive data driven discriminator based method like \cite{Shan20}, we propose a novel workflow that utilise the cues of hand and object. Specifically, we train networks predicting hand pose, hand mask and in-hand object mask to jointly predict the hand-object interaction status. We compare our method with the most recent work from Shan et al. \cite{Shan20} on selected images from EPIC-KITCHENS \cite{damen2018scaling} dataset and achieve $89\%$ accuracy on HOI (hand-object interaction) detection which is comparative to Shan's ($92\%$). However, for real-time performance, with the same machine, our method can run over $\textbf{30}$ FPS which is much efficient than Shan's ($\textbf{1}\sim\textbf{2}$ FPS). Furthermore, with our approach, we are able to segment script-less activities from where we extract the frames with the HOI status detection. We achieve $\textbf{68.2\%}$ and $\textbf{82.8\%}$ F1 score on GTEA \cite{fathi2011learning} and the UTGrasp \cite{cai2015scalable} dataset respectively which are all comparative to the SOTA methods.

\end{abstract}

\section{Introduction}
\label{sec:intro}
Detecting interaction is attractive as a way to remove redundant information in a video sequence. In egocentric perception, extracting the HOI (hand-object interaction) can be crucial in action localisation and understanding in ADLs (Activities of Daily Living) or industrial applications.


This competence is helpful within the remit of Augmented Reality (AR) and Mixed Reality (MR), too, with a variety of head-mounted devices generating large amounts of egocentric footage which needs processing to be valid. Automating information extraction from egocentric video in the form of pictorial collections or key-frames is a step in this direction. Key-frame (representative frame) extraction involves extracting the most informative frames that contain the key events in a video in terms of content \cite{lei2014novel}. Removing redundant information from long sequences in the video is critical for storage, indexing and retrieval \cite{rani2020social}.

More importantly, detecting HOI can be a new way of doing content authoring, a fundamental bottleneck problem often overlooked. Traditional content authoring techniques for AR or MR require pre-built object models and manual operation for elaborated AR/MR effects. With the extracted moment and manner of HOI task implementation, the key information can be auto edited and delivered to the user who needs guidance. HOI extraction can also impact the creation of health reports for hand disability recovery or serve to index video collections.
\begin{figure}[t]
\centering
\includegraphics[width=1\columnwidth]{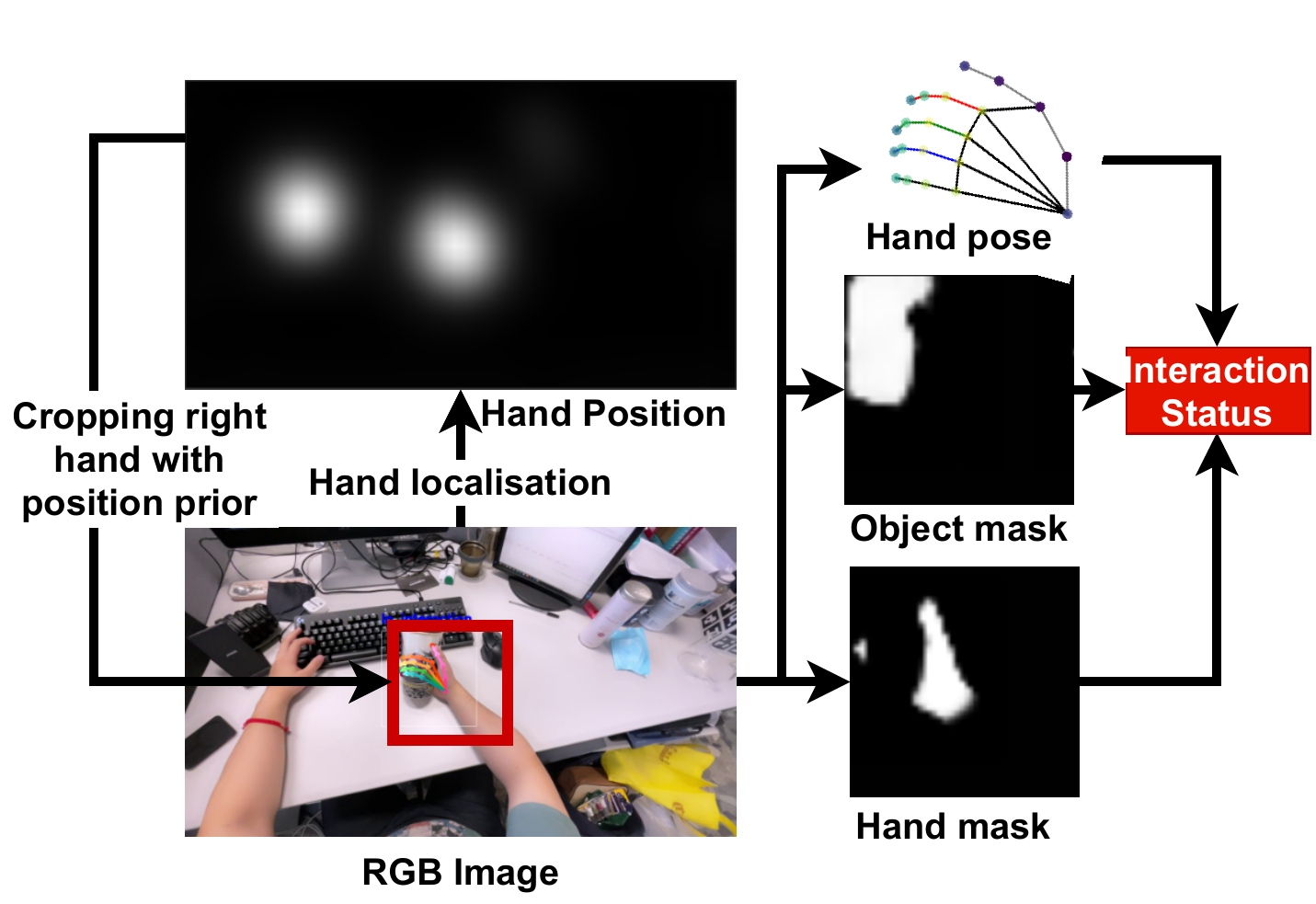} 
\caption{The pipeline of full hand-object interaction detection system.}
\label{fig:HOI_system}
\end{figure}
Unlike traditional action localisation/extraction methods, our approach does not use full-image visual features such as colour correlation, mutual information, colour moments or colour histograms, which have been widely used before \cite{rani2020social}. Instead, we focus on the human hand interacting with scene objects. In egocentric view videos, especially those tutorial videos for guidance purposes, hand-object interactions are the most important cues. Our goal is to identify the moments that convey enough information to represent hand-object interactions for non-scripted tasks. We aim for a general approach to detect and use the way hands interact with objects under actual activities without requiring the activity pre-specified or scripted.

This work proposes a workflow that is able to detect HOI (hand-object interaction) status. We use the contextual information from hand and object. Specifically, we predict the hand pose, hand mask and in-hand object mask as cues to jointly predict whether the hand is interacting/manipulating the object or not. In-hand object segmentation is a critical procedure in our method because the object is designed to be segmented when it is being manipulated. We use an FCL (fully convolutional layer) \cite{long2015fully} based network to obtain the object mask. The output is a pixel-wise possibility distribution of object position with the hand pose and the hand mask as priors.


Our contributions in this paper include: 1) Implementing the hand pose detector with a novel data collection process. 2) Annotating the data for training hand mask and in-hand object mask. 3) Implementing a hand-object interaction detection system that can be applied to any egocentric video sequence with real-time performance. 4) Evaluating our method with others on hand-object interaction detection and action segmentation.

\section{Related Work}
\label{sec:related}
The hand-object interaction detection hasn't been researched a lot independently like action recognition or action localisation, especially in egocentric perspective, people more focus on classifying the action with a data driven discriminator. While, our work aims to generalising the HOI detection for real-time applications. In this section, we mainly review the work related to egocentric interaction analysis and hand-object interaction.  
\subsection{Egocentric Analysis with HOI}
For egocentric content analysis, hands and objects all play essential roles and have been researched a lot separately or together. In Fathi et al. \cite{fathi2011understanding}, to model object interactions, a vector composed of $10$ kinds of hand-object related features is used for action classification. Pirsiavash et al. \cite{pirsiavash2012detecting} propose a model for 'active objects' for the object being interacted with human body parts and reasoning the location and size of active objects.  Li et al. \cite{li2015delving} separately model the hand pose and object with visual features and gained superior performance on activity understanding. However, solely stacking the features from the object and hand requires an extensive training model and a large amount of training data, which also has constraints with generalisation. 

Cai et al. \cite{cai2015scalable} started to explore the relationship between the semantic grasp type and egocentric activities. Later, in the work \cite{cai2016understanding}, by using the contextual information, they build models for hand grasp type and object type to jointly predict the actions. Nevertheless, we argue, embedding the hand feature to grasp type can still be dataset dependent. In the natural environment, a different in-hand object could introduce significant uncertainty on grasp type classification. To broaden the research of hand-object interaction, Shan et al. \cite{Shan20} annotated a large-scale dataset containing the hands, and the objects are being interacted with. This can be a brutal but effective way of tackling the hand-object interaction problem. While, in this work, we try to analyse hand-object interaction by further explore the higher level features that can be extracted from hands and objects.

\subsection{HOI Detection}
Detecting the hand-object interaction can also be important for an egocentric based application. Besides the object detector based method like \cite{Shan20,gkioxari2018detecting}. In work from Chen et al. \cite{chen2017}, and Lu et al. \cite{lu2019higs}, in order to find a 'touch' with the machine, the depth sensor is used to measure the distance between the hand and machine, which is effective but inflexible. Likitlersuang et al. \cite{likitlersuang2016interaction} build a hand-object interaction classifier with the optical flow and hand shape as input. The work from Schroder et al. \cite{schroder2017hand} use hand and object mask to predict the hand-object interaction status. The idea of using masks inspires our work; simultaneously, we explore the relationship between the hand-object pair and interaction with hand pose and masks. We also design our approach to be generic in that the method does not explicitly require a prior description of what objects are involved in an application to be tested.

\section{Hand Interaction Detection System}
We solve the hand-object interaction detection by utilising the contextual information of hand and object. More specifically, we use the hand pose, hand mask and object mask as cues to determine the interaction status. It can be expressed as:
$$P_{hoi} = f(H(pose), H(hand), H(object)) $$where $P_{hoi}$ is the possibility of hand-object interaction, and $H$ refers to the heatmap. The following content of the section details the implementation of the sub-tasks and how they are assembled as a system.

\subsection{Hand Pose Estimation}
Hand pose estimation has been extensively studied in the past few years. There are plenty of works detecting the hand pose on from third-person view \cite{simon2017hand,kulon2020weakly,moon2020interhand2} and first-person view \cite{mueller2018ganerated,lin2021two,mueller2017real,rogez20143d} with monocular camera. However, the researches on hand pose estimation under hand-object interaction from an egocentric view is scarce. People tend to use synthetic data to solve the occlusion issue caused by objects, but bridging the gap between the virtual data and the real environment remains unsolved. 
\subsubsection{Data Capturing}
\label{sec:pose_data}
To overcome the data shortage on HOI, we set up a multi-cam system for data capturing. Different from other multi-cam systems like \cite{simon2017hand,zimmermann2019freihand}, we use the ArUco \cite{romero2018speeded} cube for online calibration. The cube returns the 6d pose of the camera relative to the cube centre. This enables us to capture from any view without frequent system re-calibration. Notably, our system can mimic the egocentric vision for different cases (from head, chest or shoulder). 

In our setup, we use $\textbf{3}$ cameras in our system. Apparently, $\textbf{3}$ is not enough to eliminate the uncertainties caused by object occlusion. However, it is easier for the hands without objects to identify the location of joints with a well-trained 2d pose detector. By gathering all the detected 2d poses from different views, the 3d pose can be easily recovered by minimising the discrepancies observed by different cameras. Thus, we capture the hands with objects as data images, and then we remove the object while keeping hands as still as possible for ground truth data image capturing. The process is shown in figure \ref{fig:data_displaying}.

\subsubsection{Data Processing and Training}
We obtain two sets of images after the multi-cam system capturing from the previous step. Set $1$ has the images with the object in hand, and the set $2$ has the same hand pose with the object removed. For each image from set $2$, we sent it to a hand detector proposed by \cite{xiao2018simple}, and training on the datasets from the synthetic dataset GANerated Hand \cite{mueller2018ganerated}. Each view produces a set of heatmaps indicating the 2d hand joint positions $j^{2d}$. For each joint, we have a corresponding pre-posed point $j^{3d}$ in 3d space, its projections $j^{3d_proj}$ on all image planes should have minimum sum distance with the corresponding detected 2d joints $j^{2d}$. The loss can be described as:
$$Loss_{proj} =\omega_{v}^{i}\sum_{k}\sum_{i}||j^{2d}_{i}-j^{3d\_proj}_i||_2$$where $\omega_{v}^{i}$ is the $i_{th}$ joint's confidence from the 2-d joint detector of the view $v$. We use Levenberg–Marquardt algorithm to optimise the loss function. If the loss can be smaller than a threshold, we consider the data valid and take the optimised 3d joints as ground truth. The 2d joints of set $1$ (with object) can be obtained by projecting the 3d joints to the 2d plane.
\begin{figure}[t]
\centering
\includegraphics[width=1\columnwidth]{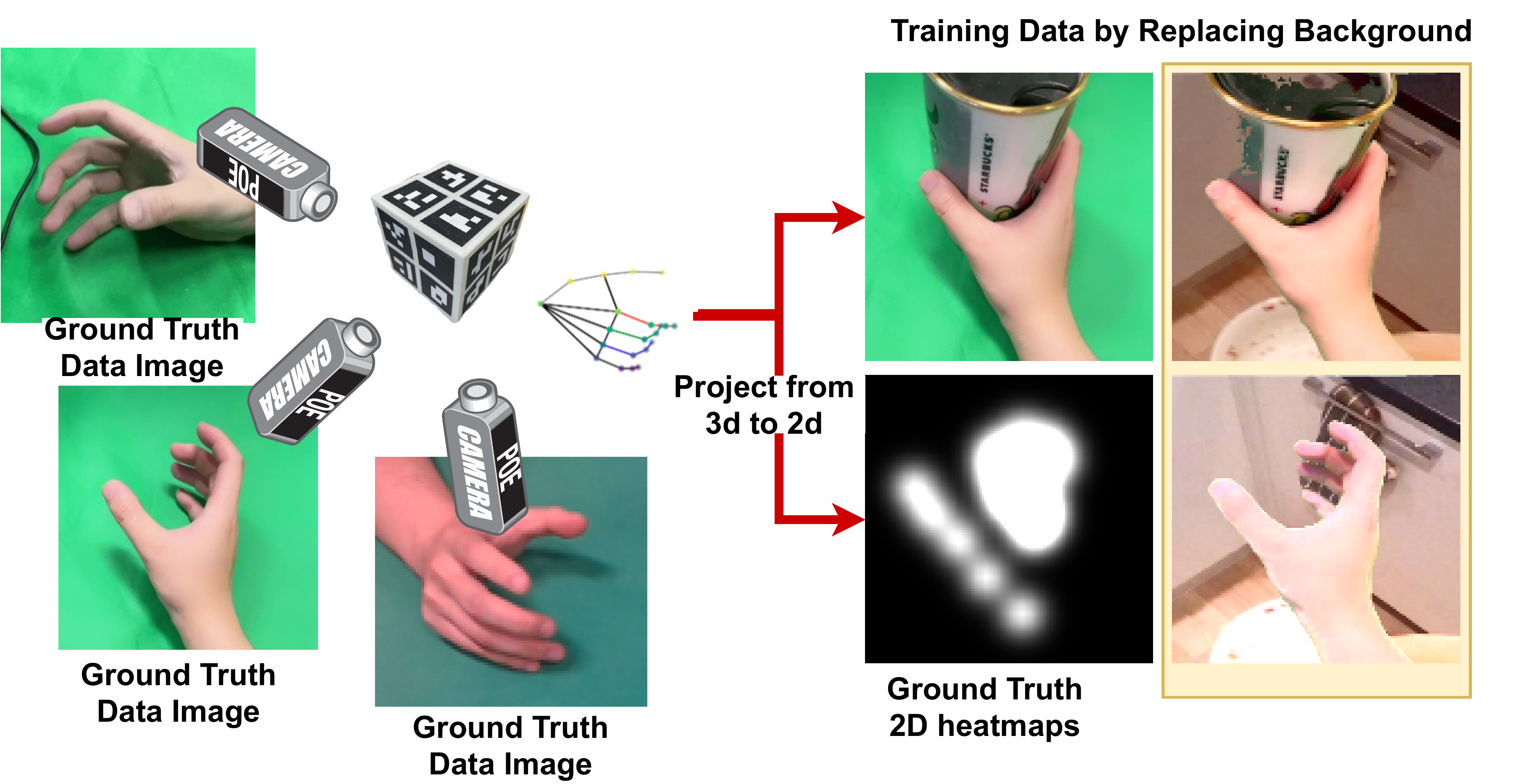} 
\caption{The figure shows the process of our data annotation. First, we capture the object-in-hand poses as the target data to label. Then, we remove all the objects and keep the hand still to capture the source of ground truth. With a green screen, we can easily replace the background with other images.}
\label{fig:data_displaying}
\end{figure}
To enrich the data augmentation, the data is collected with a green screen. As shown in figure \ref{fig:data_displaying}, with skin colour extraction in HSV colour space, the background can be randomly replaced in every training step. We have about $4k$ images that can be used for training. Same as the 2d detector used for data annotation, the hand pose detector uses an encoder-decoder paradigm inspired by the work from Xiao et al. \cite{xiao2018simple}. To prevent unreasonable poses, we add an MLP (multi-layer perceptron) as a latent hand model after the output heatmaps. The full details of the multi-cam system and the network training can be found in the supplementary material.

\subsection{In-hand Object Segmentation}
Segmenting hand and object are two topics that have been well explored separately. However, in-hand object segmentation is a complex and mostly overlooked problem of relevance in HOI (hand-object interaction) research. The main challenge could be the lack of training data and explicit applications as motivation. However, we argue, in-hand object segmentation is a strong cue for hand-object interaction analysis and can help with reducing the difficulty of in-hand object recognition for MR applications. Moreover, the contextual relationship between hand and object can be automatically encoded into the network by learning from the real hand-object interactions. As for hand-object interaction detection, this can be very helpful with identifying the non-interaction contact between hand and object.

\subsubsection{Dataset}
\begin{figure}[t]
\centering
\includegraphics[width=1\columnwidth]{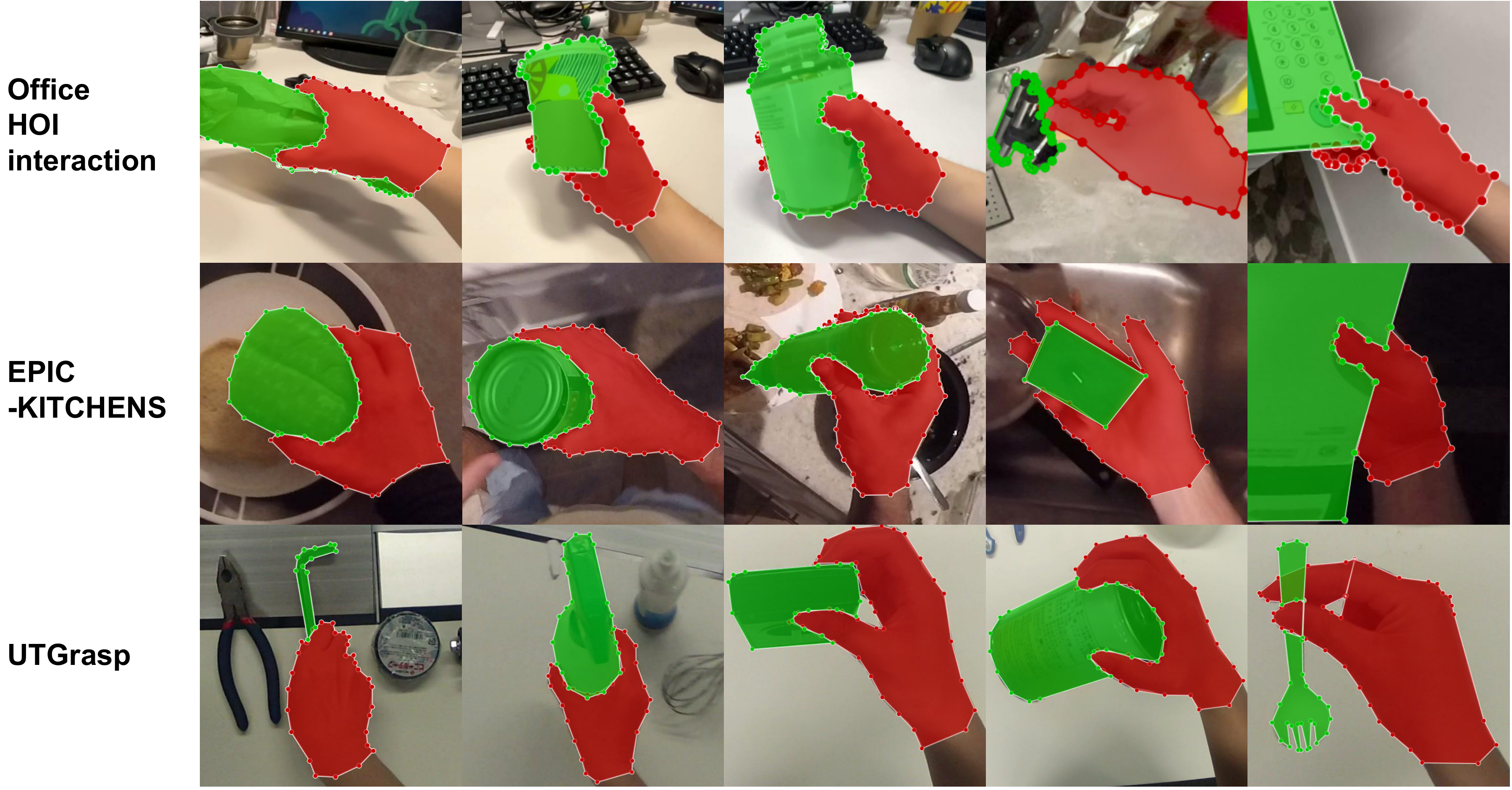} 
\caption{The figure shows the example of annotations from different datasets. The top row is our self-collected office hand-object interactions. The middle row is from EPIC-KITCHENS \cite{damen2018scaling}, and the bottom row comes from UTGrasp dataset \cite{cai2015scalable}.}
\label{fig:annotation}
\end{figure}

The dataset for training the network is composed of two parts. The first part comes from the previously labelled $4k$ images for hand pose estimation (described in section). These images are all captured in front of a green screen, we extract the background and skin colour from the HSV colour space to obtain the object mask. To have a more refined ground truth, we manually label another small dataset \textbf{GraspSeg} which contains $3k$ images from various sources, including EPIC-KITCHENS \cite{damen2018scaling}, UTGrasp \cite{cai2015scalable} and our self-collected data in office and kitchen environments. We intentionally chose images that contain different types of hand-object interactions to broaden the data variety. More importantly, we also choose the hands without hand-object interaction or non-interaction contacting as for negative samples. All the data images are annotated with the hand mask and in-hand object mask. Examples of our annotation are shown in figure \ref{fig:annotation}. 

\subsubsection{Grasp Response Map and Interaction Detection}
\begin{figure*}[t]
\centering
\includegraphics[width=2\columnwidth]{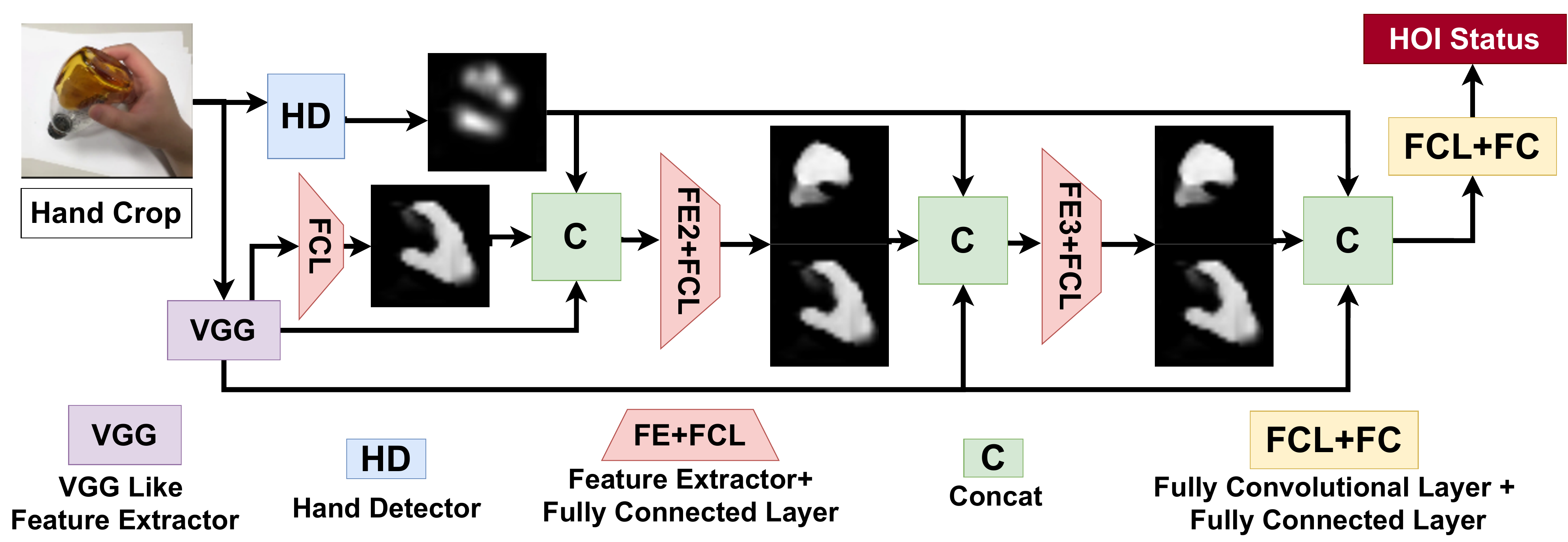} 
\caption{The picture shows the schematic of our network. We use a 3-layer cascade structure to feed hand pose cues and hand mask cues to predict the 'Grasp Response Map (object mask)' and the interaction status. }
\label{fig:GRM}
\end{figure*}

The network structure we build to generate masks is shown in figure \ref{fig:GRM}. The network is cascaded structured, and it contains three stages. The first stage predicts the hand mask with FCL (fully convolutional layer) \cite{long2015fully}. The input for the second stage is concatenated by the predicted hand mask, hand pose heatmaps and extracted features from the backbone. The object masks $m_{obj}$ is produced together with the hand mask from the stage $2$ and $3$. It can be regarded as a possible distribution under the joint condition of hand pose $H_{k(p)}$, hand mask $M_{hand}$ and the corresponding extracted features. We call the object mask $M_{obj}$ 'Grasp Response Map' (GRM) because the specific combination of hand pose determines its response, and hand mask and the objects' visual appearance. 

The hand cues and object cues jointly determine the hand-object interaction status. In our network design, we add a small, fully convolutional layer to reduce the dimension of the features and predict the possibility of hand-object interaction with a fully connected layer head. 

\subsection{Experiments and Results}
In this section, we mainly evaluate in-hand object segmentation and the performance of our HOI (hand-object interaction) detector quantitatively and visually. Testing data is chosen from EPIC-KITCHENS \cite{damen2018scaling}, GTEA \cite{fathi2011learning} and UTGrasp \cite{cai2015scalable}. Evaluating the hand pose detector is not within the scope of this paper. It can be found in supplementary materials.

\subsubsection{In-Hand Object Segmentation}
\begin{figure*}[tb]
 \centering 
 \includegraphics[width=2\columnwidth]{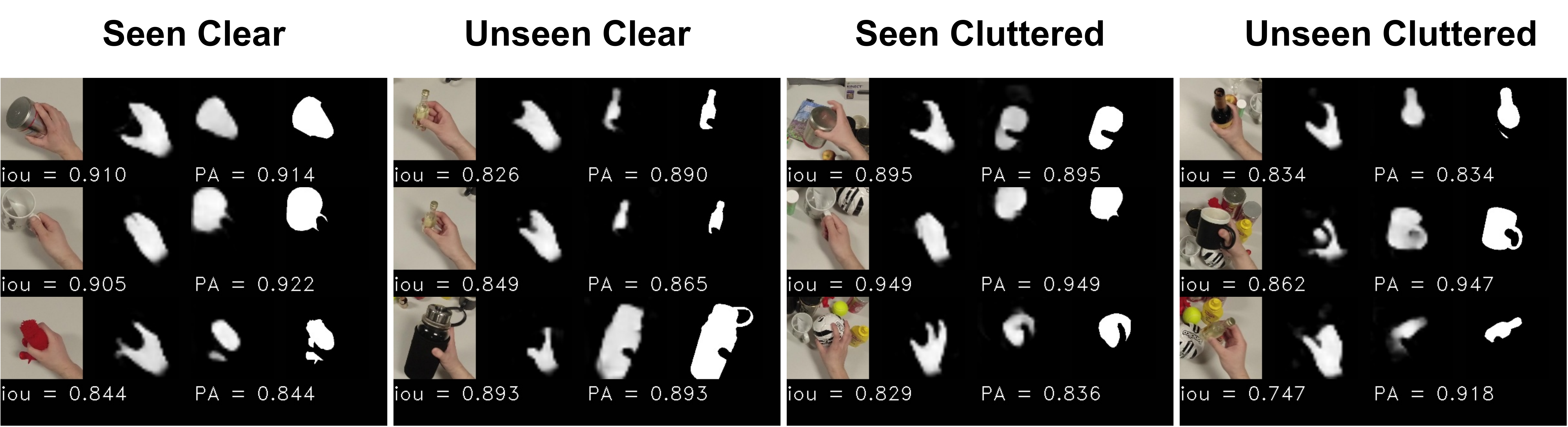}
 \caption{The figure shows the example of testing results on different conditions. In each set, from left to right are input hand crop, hand segmentation result, object segmentation result and ground truth object mask. 'seen'/'unseen': whether the object is included in the dataset. 'clear/cluttered': whether the background is clear or not. 'IOU': intersect over union between predicted object mask and ground truth object mask. 'PA': pixel accuracy.}
 \label{fig:segmentation_rst}
\end{figure*}

To evaluate the performance of our in-hand object segmentation, We additionally label $400$ masks of in-hand objects categorised into 'seen clear', 'unseen clear', 'seen cluttered' and 'unseen cluttered'. Where 'seen'/'unseen' indicates whether the object is included in the dataset and 'clear/cluttered' means whether the background is clear or cluttered. The quantitative results over the $400$ test images are shown in table \ref{tab:rst} and the visual results are shown in figure \ref{fig:segmentation_rst}. Our network can segment the in-hand object in different situations. Even for the unseen objects with a cluttered background, we achieve $0.64$ on intersect over union and $0.74$ on pixel accuracy. 

\begin{table}
\begin{center}
\begin{tabular}{|l|c|c|c|c|c}
\hline
  & S Clear & U Clear & S Cluttered & U Cluttered \\
\hline\hline
IOU & \textbf{0.84} & 0.77& 0.78& 0.64 \\
PA & \textbf{0.86} & 0.79& 0.85& 0.74 \\
\hline
\end{tabular}
\end{center}
\caption{The table shows the quantitative results of object segmentation under different conditions. 'S' stands for 'seen' and 'U' stands for 'unseen'.}
\label{tab:rst}
\end{table}

\subsubsection{Test Set for HOI Detection}
\begin{figure}[t]
\centering
\includegraphics[width=1\columnwidth]{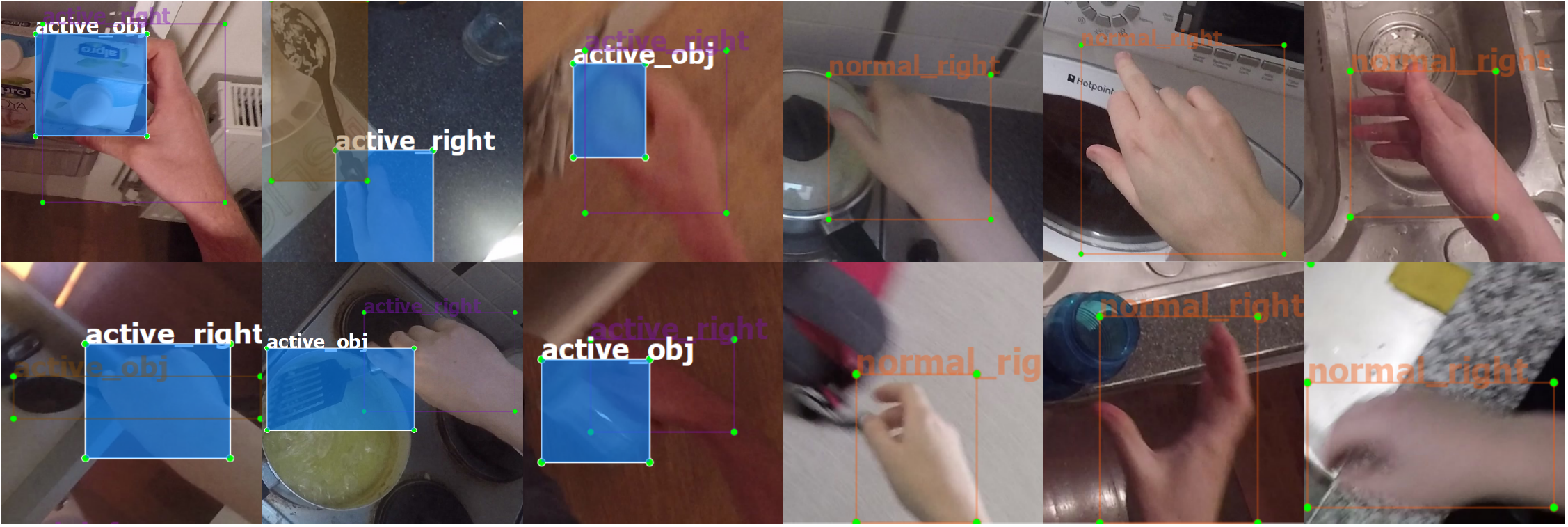} 
\caption{The figure shows the examples of testing data we re-labelled from EPIC-KITCHENS\cite{damen2018scaling}.}
\label{fig:test_data}
\end{figure}

As far as we know, there is no widely accepted dataset for egocentric hand-object interaction detection. To facilitate the evaluation, we create a small test set from the EPIC-KITCHENS \cite{damen2018scaling} dataset. The EPIC-KITCHENS contains a large amount of unscripted hand-object interactions. We randomly select $3k+$ frames from the images officially extracted from the videos, and the frames they provided have been auto labelled by the detector from the work of Shan \cite{Shan20}. To obtain the ground truth, we manually check and correct the auto labelled bounding boxes and interaction status. The images that do not have a hand or cannot be detected by the model provided by \cite{Shan20} are not considered. The example re-labelled images are demonstrated in figure \ref{fig:test_data}. 

\subsubsection{HOI Detection on Selected Frames}
\label{sec:HOI_DETECTION}
In this experiment, we mainly evaluate:
\begin{itemize}
    \item Our proposed method. We report the results of our proposed method on the test set images.
    \item Shan's model. We report the results detected by Shan's model \cite{Shan20}.
    \item The role of hand pose. We train our network without the cue of hand pose heatmaps and report the performance of the trained detector. 
    \item VGG based image classification. We use a pre-trained VGG as backbone to classify the hand-object interaction status.
\end{itemize}

\begin{table}
\begin{center}
\begin{tabular}{|l|c|c|c|c|c}
\hline
  & Ours & Shan et al. & No hand pose & VGG based \\
\hline\hline
Acc & 0.89& \textbf{0.92}& 0.79& 0.68 \\
\hline
\end{tabular}
\end{center}
\caption{The table shows the quantitative results on the test set.}
\label{tab:rst1}
\end{table}
In table \ref{tab:rst1}, we report the results from different experiment setups. We found the classifier only use VGG as backbone has the worst performance; it is just $\textbf{18\%}$ better than random guess ($\textbf{0.5}$). Moreover, the result of the 'No hand pose' setup proves the necessity of adding a hand pose cue in our model. Our method achieves $\textbf{0.89}$ accuracy on hand-object interaction detection based on the test set. It is comparable to the results from Shan's \cite{Shan20} model. Nevertheless, the size of our model is less than $\textbf{100M}$ (hand pose detector: $\textbf{84}$, and HOI detector: $\textbf{14}$) which is much small than Shan's ($\textbf{361M}$). As for real-time performance, compared with about $\textbf{1}\sim\textbf{2}$ FPS on Shan's model, our method can run at over $\textbf{30}$ FPS on an old machine with Nvidia Quadro M2000 GPU (4GB) and i5 processor. 

\begin{figure*}[t]
\centering
\includegraphics[width=2\columnwidth]{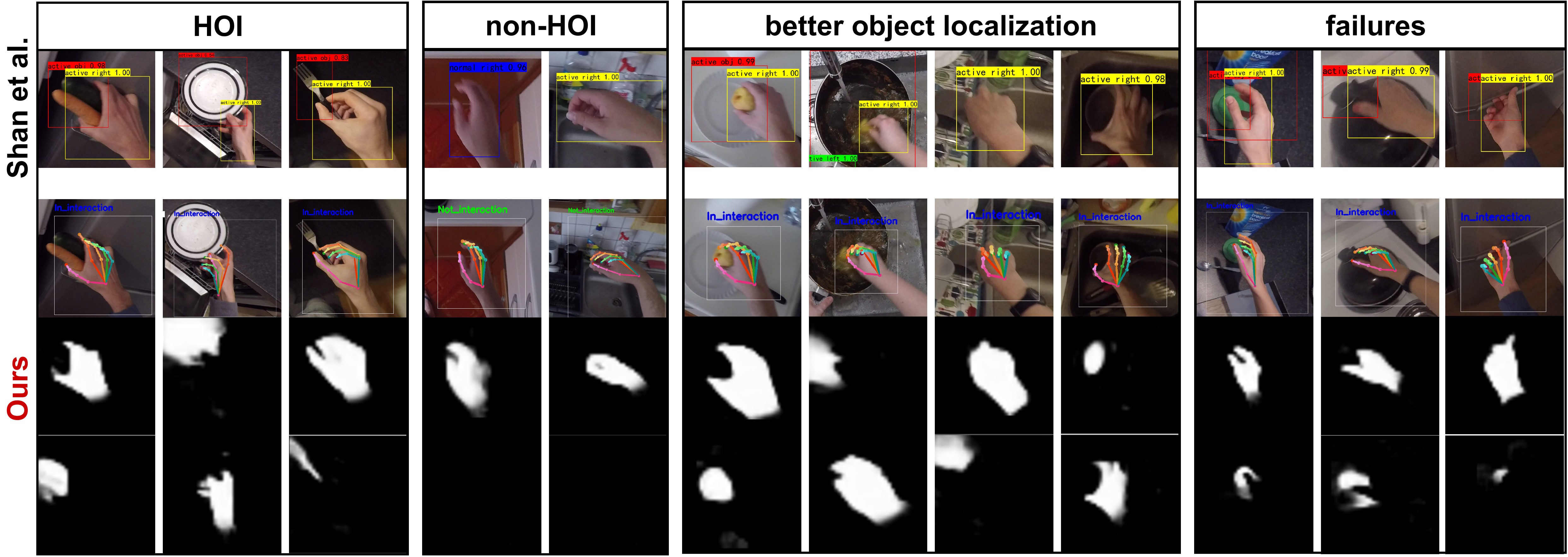} 
\caption{The figure shows the visual comparison between our method and Shan's \cite{Shan20}. Our method tends to be good at locating HOI with a relatively small object. While Shan's model has better performance on HOI detection with a larger object.}
\label{fig:visual_results}
\end{figure*}

The visual results are shown in figure \ref{fig:visual_results}. Both our method and Shan's \cite{Shan20} model perform well on HOI detection. Our method has better performance on locating the same object being manipulated. Because our method is performed with the cropped hand area, locating the object partially visible in the crop is an apparent shortage.

\begin{figure*}
\centering
\includegraphics[width=2\columnwidth]{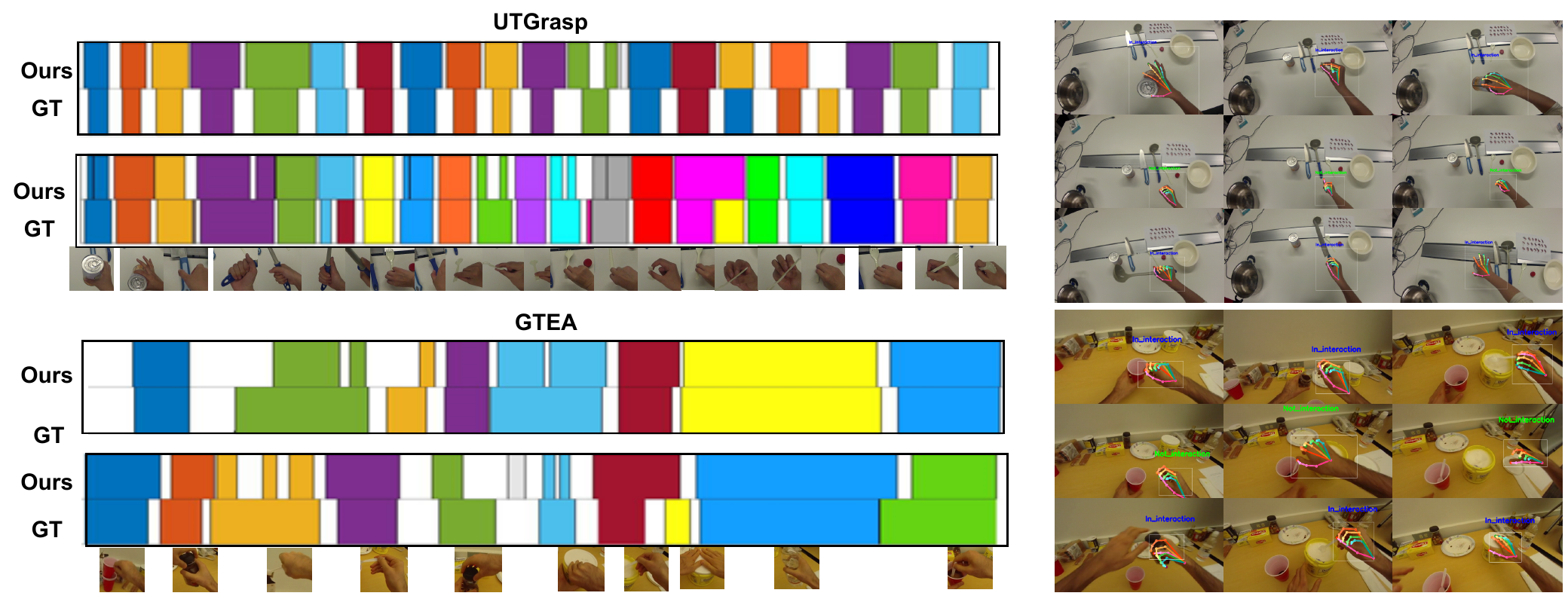} 
\caption{The figure shows the examples of video segmentation by our HOI system. The left part shows the two example segmentation from each dataset. The right part shows some visual results of our HOI detector.}
\label{fig:seg_rst}
\end{figure*}
\subsection{HOI Video Extraction}
To further demonstrate the capability of our HOI detector in the video sequence. We build a system that can automatically extract the HOI segments from the egocentric video. However, the egocentric dataset like EPIC-KITCHENS \cite{damen2018scaling} contains a large amount of HOI annotations without a clear hand in the scene, which make evaluation pointless. While, The prescript dataset GTEA \cite{fathi2011learning} and UTGrasp \cite{cai2015scalable} are performed in front of a table without large-scale ego-motion. This is closer to the ideal condition of delivering guidance with XR technology, which is used for evaluating the HOI video segmentation.

\subsubsection{Hand Interaction System}
We implement a system that contains hand localisation, hand pose tracking and hand-object interaction detection. (shown in figure \ref{fig:HOI_system}) For simplicity, the system is designed for working with a single right hand. To \textbf{localise the hand}, we resize the whole frame to $48\times28$ and regard the hand area as a key point. We train a fully convolutional network that outputs a heatmap showing the possible location of hands and their ID (left or right). The ground truth is obtained by putting a Gaussian distribution in the centre of bounding boxes regardless of the original box size. The process can be interpreted as the ROI (region of interest) extraction with a heatmap. \textbf{Hand ID classification} is designed with $3$ kinds of outputs: 'left hand', 'right hand' and 'two hands'. When 'two hands' are detected, we simply take the one on the right as the right hand. A crop around the detected 'keypoint' is sent to estimate the hand pose, and the \textbf{tracking} is accomplished by re-centring the tracking box to the centre of hand joints over the successive frames.

\subsubsection{Experiments and Results}
To obtain the results of HOI detection, we simply apply our HOI system on the videos for evaluation and record the HOI status frame-wisely. For the GTEA \cite{fathi2011learning} dataset, two hands are involved in most of the interactions. Because the camera wearers are all right-handed, we take the HOI status of the right hand as a result. Both GTEA \cite{fathi2011learning}, and UTGrasp provides the time stamps for actions or object interactions. To facilitate evaluation, we report frame accuracy and F1 score of IOU at $50\%$, which is the same as the evaluation protocol used in Farha et al. \cite{farha2019ms}. We compare the segmentation results with a video-based method MS-TCN \cite{farha2019ms} and Shan's \cite{Shan20} frame-based model. To overcome the noise (false detection) problem of Shan's and our frame-based method, we apply a temporal filter with half-second to smooth the detection results over time. Figure \ref{fig:seg_rst} shows the example results of our HOI system.

\begin{table}
\centering
\caption{The $F1$ score with overlapping thresholds (IOU) $50\%$ and frmae accuracy on GTEA \cite{fathi2011learning} dataset. MS-TCN: Multi-stage temporal convolution network. \cite{farha2019ms}. Shan's: Results obtained by running the model provided by Shan et al. \cite{Shan20}.}
\begin{tabular}{ |l|c|c| } 

\hline
     Dataset &UTGrasp & GTEA  \\
\hline
     Ours F1@$50\%$&\textbf{ 89.3}\% & 68.2\% \\
\hline
      MS-TCN  F1@$50\%$&-&\textbf{ 69.8}\%\\ 
\hline
     Shan's F1@$50\%$& 82.1\% & 62.4\%\\ 
\hline
\hline

     Ours Acc& 82.8\% & 71.5\% \\
\hline
      MS-TCN Acc& - &\textbf{76.3}\%\\ 
\hline
     Shan's Acc& \textbf{83.7}\% & 65.6\%\\ 
\hline

\end{tabular}
\label{tab:seg_rst}
\end{table}
We report our results in table \ref{tab:seg_rst}. For UTGrasp dataset, we achieve $89.3\%$ F1 scores on video action segmentation and $82.8\%$ on HOI status accuracy. Our system outperforms the Shan et al. \cite{Shan20} on the action segmentation and has a comparative result on frame-wise HOI status accuracy. For GTEA \cite{fathi2011learning} dataset, our result is very close to the MS-TCN's \cite{farha2019ms}. MS-TCN is a SOTA method of action segmentation that uses the whole video feature to predict the segments.

For the running time, with the same machine (Nvidia Quadro M2000 GPU (4GB) and i5 processor), our method can be applied in real-time, the Shan's model is about $\textbf{1}\sim\textbf{2}$ FPS. The video feature-based method takes longer for data pre-processing (like producing optical flow by using other networks). 

\section{Conclusion}
We have presented a work that can detect hand-object interactions from an egocentric perspective. To achieve the goal, we implemented sub-tasks, including hand pose estimation and hand-object pair segmentation. We use a novel data labelling approach for hand pose estimation to tackle the in-hand object occlusion problem and achieve good performance with training on a small amount of data. Also, we annotated a dataset GraspSeg and trained with our novel network for hand and in-hand object mask prediction. We predict the hand-object interaction status with detected hand pose and masks. Our method is evaluated and compared with the SOTA on selected frames from EPIC-KITCHENS \cite{damen2018scaling}, and GTEA \cite{fathi2011learning}, UTGrasp \cite{cai2015scalable} in form of video segmentation.

{\small
\bibliographystyle{ieee_fullname}
\bibliography{egbib}
}

\end{document}